\documentclass[runningheads]{llncs}
\usepackage{graphicx, subfig}
\usepackage[font=small]{caption}
\usepackage{amsmath,amssymb} 

\usepackage{color,algorithm,algpseudocode}
\usepackage{subfig}
\usepackage{array}
\usepackage{multirow}

\begin{document}

\title{PCA-RECT: An Energy-efficient Object Detection Approach for Event Cameras} 
\titlerunning{PCA-RECT for Event-based Object Detection} 


\author{Bharath Ramesh*\orcidID{0000-0001-8230-3803} \and
Andr\'{e}s Ussa \and
Luca Della Vedova \and 
Hong Yang \and
Garrick Orchard}
%

\authorrunning{B. Ramesh et al.} 


\institute{Temasek Laboratories, National University of Singapore, Singapore 117411 
\email{bharath.ramesh03@u.nus.edu}\\
\url{http://sites.google.com/view/bharath-ramesh/} }

\maketitle

\begin{abstract}
We present the first purely event-based, energy-efficient approach for object detection and categorization using an event camera. Compared to traditional frame-based cameras, choosing event cameras results in high temporal resolution (order of microseconds), low power consumption (few hundred mW) and wide dynamic range (120 dB) as attractive properties. However, event-based object recognition systems are far behind their frame-based counterparts in terms of accuracy. To this end, this paper presents an event-based feature extraction method devised by accumulating local activity across the image frame and then applying principal component analysis (PCA) to the normalized neighborhood region. Subsequently, we propose a backtracking-free \emph{k}-d tree mechanism for efficient feature matching by taking advantage of the low-dimensionality of the feature representation. Additionally, the proposed \emph{k}-d tree mechanism allows for feature selection to obtain a lower-dimensional dictionary representation when hardware resources are limited to implement dimensionality reduction. Consequently, the proposed system can be realized on a field-programmable gate array (FPGA) device leading to high performance over resource ratio. The proposed system is tested on real-world event-based datasets for object categorization, showing superior classification performance and relevance to state-of-the-art algorithms. Additionally, we verified the object detection method and real-time FPGA performance in lab settings under non-controlled illumination conditions with limited training data and ground truth annotations.

\end{abstract}
\section{Introduction}
\label{sec:intro}
Through these fruitful decades of computer vision research, we have taken huge strides in solving specific object recognition tasks, such as classification systems for automated assembly line inspection, hand-written character recognition in mail sorting machines, bill inspection in automated teller machines, to name a few. Despite these successful applications, generalizing object appearance, even under moderately controlled sensing environments, for robust and practical solutions for industrial challenges like robot navigation and sense-making is a major challenge. This paper focuses on the industrially relevant problem of real-time, low-power object detection using an asynchronous event-based camera \cite{Brandli2014} with limited training data under unconstrained lighting conditions. Compared to traditional frame-based cameras, event cameras do not have a global shutter or a clock that determines its output. Instead, each pixel responds independently to temporal changes with a latency ranging from a low of tens of microseconds to a high of few milliseconds. This local sensing paradigm naturally results in a wider dynamic range (120 dB), as opposed to the usual 60 dB for frame-based cameras.  
\par
Most significantly, event cameras do not output pixel intensities, but only a spike output with a precise timestamp, also termed an event, that signifies a sufficient change in log-intensity of the pixel. As a result, event cameras require lower transmission bandwidth and consume only a few hundred~mW vs. a few~W by standard cameras \cite{Posch2014}. In summary, event-based cameras offer a fundamentally different perspective to visual imaging while having a strong emphasis on low-latency and low-power algorithms \cite{Kueng2016,Conradt2009,Ni2012,Delbruck2013}. 
\par
Despite the notable advantages of event cameras, there still remains a significant performance gap between event camera algorithms and frame-based counterparts for various vision problems. This is partly due to a requirement of totally new event-by-event processing paradigms. However, the burgeoning interest in event-based classification/detection is focused on closing the gap using deep spiking neural networks \cite{OConnor2013,Lee2016}, something that again entails dependence on powerful hardware like its frame-based counterpart. On the other hand, a succession of frames captured at a constant rate (say 30 Hz), regardless of the scene dynamics and ego-motion, works well with controlled scene condition and camera motion. Frame-based computer vision algorithms have benefited immensely from sophisticated methodologies that reduce the computational burden by selecting and processing only informative regions/keypoints within an image \cite{lowe2004distinctive,Galleguillos2008,Ramesh2017c,Vikram2012}. In addition, frame-based sensing has led to high hardware complexity, such as powerful GPU requirements for state-of-the-art object detection frameworks using deep neural networks \cite{Redmon2018,Ren2017}.
\par
\begin{figure}[t]
	\centering
	\includegraphics[width=12.5cm]{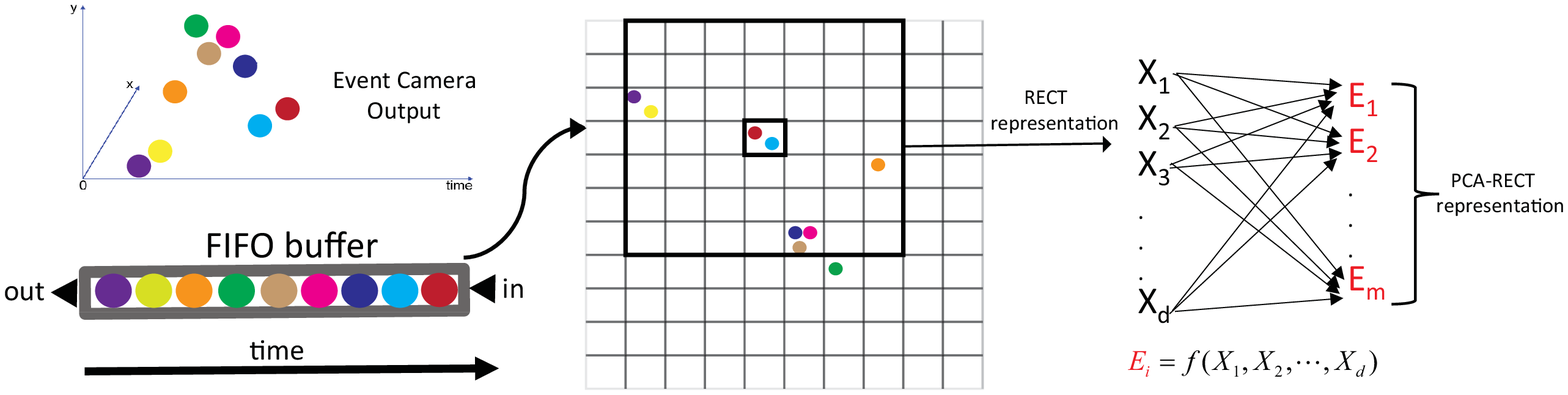}
	\caption{PCA-RECT representation (best viewed on monitor). Useful events are sub-sampled and filtered after applying nearest-neighbor temporal filtering and refractory filtering, termed as rectangular event context transform (RECT). The sparser RECT event representation is updated dynamically using a first in, first out (FIFO) buffer. Subsequent feature extraction is carried out by applying principal component analysis (PCA) to project RECT onto a lower-dimensional subspace to obtain the final PCA-RECT feature representation}
	\label{fig:intro}
\end{figure}

In contrast to the above works, this paper introduces a simple, energy-efficient approach for object detection and categorization. Fig.~\ref{fig:intro} illustrates the local event-based feature extraction pipeline that is used for classification using a dictionary-based method. Accordingly, efficient feature matching with the dictionary is required, which is handled by a backtracking-free branch-and-bound \emph{k}-d tree. This proposed system was ported to a field programmable gate array (FPGA) with certain critical design decisions, one of which demanded a virtual dimensionality reduction method based on the \emph{k}-d tree, to accommodate very low-power computational needs.


\section{Event Cameras}
For real-time experiments, we use the commercial event camera, the Dynamic and Active-pixel Vision Sensor (DAVIS) \cite{Brandli2014}. It has 240 $\times$ 180 resolution, 130 dB dynamic range and 3 microsecond latency. The DAVIS can concurrently output a stream of events and frame-based intensity read-outs using the same pixel array. An event consists of a pixel location ($x$, $y$), a binary polarity value ($p$) for positive or negative change in log intensity and a timestamp in microseconds ($t$). In this work, polarity of the events are not considered, and only the event stream of the DAVIS is used. 

\subsection{Related Work}
Since event-based vision is relatively new, only a limited amount of work addresses object detection using these devices \cite{Liu2016,Lenz2018}. Liu \emph{et al}. \cite{Liu2016} focuses on combining a frame-based CNN detector to facilitate the event-based module. We argue that works using deep neural networks for event-based object detection may achieve good performance with lots of training data and computing power, but they go against the idea of low-latency, low-power event-based vision. In contrast, \cite{Lenz2018} presents a practical event-based approach to face detection by looking for pairs of blinking eyes. While \cite{Lenz2018} is applicable to human faces in the presence of activity, we develop a general purpose event-based, object detection method using a simple feature representation based on local event aggregation. Thus, this paper is similar in spirit to the recently spawned ideas of generating event-based descriptors, such as histogram of averaged time surfaces \cite{Sironi2018} and log-polar grids \cite{Ramesh2017d,Ramesh2017a}. Moreover, the proposed object detection and categorization method was accommodated on FPGA to demonstrate energy-efficient low-power vision.
\section{Method}
\label{sec:method}
We follow the event-based classification framework proposed in \cite{Ramesh2017a}, with the following crucial changes: a new descriptor (PCA-RECT), a virtual dimensionality reduction technique using \emph{k}-d trees (vPCA) and a simplified feature matching mechanism to account for hardware limitations. The framework \cite{Ramesh2017a} consists of four main stages: feature extraction, feature matching with a dictionary, dictionary representation followed by a linear classifier. Additionally, we incorporate an object detector in the framework as explained in the following subsections.
\subsection{PCA-RECT}
\label{sec:pcarect}
Each incoming event, ${\bf{e}}_i = (x_i,y_i,t_i,p_i)^T$ with pixel location $x_i$ and $y_i$, timestamp $t_i$, polarity $p_i$, is encoded as a feature vector ${\bf{x}}_i$. To deal with hardware-level noise from the event camera, two main steps are used: (1) nearest neighbour filtering and (2) refractory filtering. We define a spatial Euclidean distance between events as, 
\begin{equation}
{D_{i,j}} = \left| {\left| {\left( {\begin{array}{*{20}{c}}
{{x_i}}\\
{{y_i}}
\end{array}} \right) - \left( {\begin{array}{*{20}{c}}
{{x_j}}\\
{{y_j}}
\end{array}} \right)} \right|} \right| ~.
\label{eq:distanceevents}
\end{equation}
Using the above distance measure, for any event we can define a set of previous events within a spatial neighborhood, $N\left( {{\bf{e}}_i,\gamma } \right) = {\rm{\{ }}{\bf{e}}_i\;{\rm{|}}\;j < i,\;{D_{i,j}} < \gamma \} {\rm{\;}}$, where $\gamma = \sqrt{2}$ for an eight-connected pixel neighbourhood. When the time difference between the current event and the most recent neighboring event is less than a threshold, ${{\bf{\Theta }}_{noise}}$, the filter can be written as
\begin{equation}
{F_{noise}}\left( {\bf{e}} \right) = {\rm{\{ }}{\bf{e}}_i{\rm{|}}\;N( {{\bf{e}}_i,\;\sqrt 2 } )\backslash N( {{\bf{e}}_i,\;0} )\; \ni \;{\bf{e}}_j\;{\rm{|}}\;{t_i} - {t_j} < {{\bf{\Theta }}_{noise}}{\rm{\} }}~.
\label{eq:noisefilter}
\end{equation}
When the neighborhood is only the current pixel, $\gamma = 0$, the set of events getting through the refractory filter ${F_{ref}}$ are those such that,
\begin{equation}
{F_{ref}}\left( {\bf{e}} \right) = {\rm{\{ }}{\bf{e}}_j{\rm{|}}\;\;{t_i} - {t_j} > {{\rm{\Theta }}_{ref}}\;\forall \;j\;|\;{\bf{e}}_j \in N\left( {{\bf{e}}_j,\;0} \right)\}~. 
\label{eq:reffilter}
\end{equation}
Cascading the filters, we can write the filtered incoming events as,
\begin{equation}
\left\{ \hat{\bf{e}} \right\} = \;{F_{noise}}\left( {\;{F_{ref}}\left( {\bf{e}} \right)\;} \right) ~.
\label{eq:filtered}
\end{equation}

As shown in Fig.~\ref{fig:intro}, the incoming events $\hat{\bf{e}}_i$ are first pushed into a FIFO buffer. The FIFO queue is then used to update an event-count matrix $C \in \mathbb{R}^{m \times n}$, where $m$ and $n$ denote the number of rows and columns of the event camera output.
\begin{equation}
C(x_{i},y_{i}) = C(x_{i},y_{i}) + 1 ~.
\label{eq:queue1}
\end{equation}
Before pushing the latest event, the FIFO buffer of size s is popped to make space and simultaneously update the count matrix C,
\begin{equation}
C(x_{i-s},y_{i-s}) = C(x_{i-s},y_{i-s}) - 1 ~.
\label{eq:queue2}
\end{equation}
The event-count $C$ is pooled to build local representations, which are further aggregated to obtain the RECT representation of each event. In particular, let $A$ be a $p\times p$ square filter, the 2-D convolution is defined as,
\begin{equation}
\label{equ1}
   R(j,k) = \sum\limits_p {\sum\limits_q {A(p,q)C(j - p + 1,k - q + 1)} } ~,
\end{equation}
where p run over all values that lead to legal subscripts of $A(p,p)$ and $C(j-p+1,k-p+1)$. In this work, we consider a filter containing equal weights (commonly known as an averaging filter) for simplicity, while it is worth exploring Gaussian-type filters that can suppress noisy events. The resultant 2-D representation is termed as filtered matrix $R \in \mathbb{R}^{(m/p) \times (n/p)}$, where the filter dimensions are chosen to be give integer values for $m/p$ and $n/q$ or conversely $C$ is zero-padded sufficiently. Subsequently, the RECT representation for $\hat{\bf{e}}_i$ is obtained as a patch ${\bf{u}}_i$ of dimension $d$ centered at $R(y/p,x/p)$. Subsequently, the filtered event-count patch is projected on-to a lower-dimensional subspace using principal component analysis (PCA) for eliminating noisy dimensions and improving classifier accuracy.

\subsection{Feature Selection and Matching using \emph{K}-d Trees}
\label{sec:kdtreefast}
The PCA-RECT feature representation for each event is classified using a dictionary type method \cite{Ramesh2017a} that can handle the recognition of the desired object categories. However, exhaustive search is too costly for nearest neighbor matching with a dictionary, and approximate algorithms can be orders of magnitude faster than exact search, while almost achieving par accuracy. 
\par
In the vision community, \emph{k}-d tree nearest-neighbor search is popular \cite{Silpa-Anan2008,Muja2009}, as a means of searching for feature vectors in a large training database. Given \emph{n} feature vectors ${\bf{x}}_i \in \mathbb{R}^{d'}$, the \emph{k}-d tree construction algorithm recursively partitions the $d'$-dimensional Euclidean space into hyper-rectangles along the dimension of maximum variance. However, for high dimensional data, backtracking through the tree to find the optimal solution takes a lot of time.  
\par
This paper proposes a simple, backtracking-free branch-and-bound search for dictionary matching, taking advantage of the low-dimensionality of the PCA-RECT representation. The hypothesis is that, in general, the point recovered from the leaf node is a good approximation to the nearest neighbor in low-dimensional spaces, and performance degrades rapidly with increase in dimensionality, as inferred from the intermediate results in \cite{Beis1997}. In other words, with ${(\log _2}~n) - 1$ scalar comparisons, nearest neighbor matching is accomplished without an explicit distance calculation. While the PCA-RECT representation is useful for software implementations, an extra PCA projection step can be computationally demanding on FPGA devices. To this end, we propose a virtual PCA-RECT representation based on the \emph{k}-d tree, termed as vPCA-RECT. 
\par
\subsubsection{vPCA-RECT}
\label{sec:pcarectvirtual}
A key insight is that only a fraction of the data dimensions are used to partition the \emph{k}-d tree, especially when the dictionary size is only a few times more than the feature dimension. Therefore, instead of using the PCA-RECT representation, an alternative dimensionality reduction scheme can be implemented by discarding the unused dimensions in the \emph{k}-d tree structure. In other words, the RECT representation is first used to build a \emph{k}-d tree that selects the important dimensions (projection $\pi$), which are then utilized for dictionary learning and classification. It is worth noting that exactly the same \emph{k}-d tree will be obtained if the RECT data is first projected by $\pi$ onto a subspace that is aligned with the coordinate axes. Since no actual projection takes place, we refer to this as a virtual projection -- the irrelevant dimensions chosen by the \emph{k}-d tree are discarded to obtain a lower-dimensional feature representation.

\subsection{Event-based Object Categorization and Detection}
\label{sec:categdetetmethods}
\emph{The learning stage}: Using either the PCA-RECT or vPCA-RECT event representation, the learning process corresponds to creating a set of $K$ features denoted as $M= \{1,2,\cdots,K\}$ to form the dictionary. First, a simple sampling process is carried out such that, during training, a large pool of event representations of various categories and at random positions are extracted from a target set of events. In our setup, the dictionary features are learned from the sampled training set using clustering for all the objects jointly. 
\par
The learning stage for detection builds on top of the categorization module, in such a way that the learning process corresponds to selecting a subset of features from the dictionary for each object. In contrast to the learning phase of the categorization module, the detector features are selected from the whole training set in a supervised one-vs-all manner. 
\par
We propose to evaluate the balanced matches $Y_+^k$ to each dictionary feature $f_k$ from the target events against the matches $Y_-^k$ for all the other events to the respective feature. Mathematically, the ratio   
\begin{eqnarray}
D(k) = \frac{\beta_+^k Y_+^k}{\beta_-^k Y_-^k}~,  ~\text{where}~
\beta_+^k = \frac{|Y_+^k|}{\sum\limits_{k = 1}^K |Y_+^k|}, ~\text{and}~
\beta_-^k = \frac{|Y_-^k|}{\sum\limits_{k = 1}^K |Y_-^k|} ~,
\end{eqnarray}
is to be maximized. The balancing component $\beta_+^k$ denotes the percentage of target events matched to the dictionary feature $f_k$. Similarly, $\beta_-^k$ denotes the percentage of non-target events matched to the dictionary feature $f_k$. Thus, choosing the detector features with the D--largest ratios completes the learning phase. 
\noindent\par
\emph{The classification/detection stage}: At runtime, the event representations are propagated through the \emph{k}-d tree. On the one hand, the distribution of the dictionary features are then extracted and further passed to a simple linear classifier (we experimented with both linear SVM and Kernel Methods). On the other hand, the event representations propagated through the \emph{k}-d tree are matched with the detector features . Those matched events are used to update a location map for the target object and the region with the highest activation is considered to be the final detection result. 

\section{FPGA Implementation}
\label{sec:fpgadetails}
\subsection{Categorization Pipeline}
\label{subsec:fpgaclassification}
In order to showcase energy-efficient event-based object recognition, the FPGA implementation of the algorithm is designed as a series of four independent hardware units: event sub-sampling, vPCA-RECT generation, a recursive \emph{k}-d tree and a SVM classifier output on an event-by-event basis, each of which has an independent block design. Generally, these hardware counterparts are not a direct application of the algorithm presented in the earlier section, i.e., certain design decisions were taken for this task, among them, to desist the use of an extra PCA projection along the pipeline. 
\par
\begin{figure}[t]
	\centering
\subfloat[Sub-sampling module]{\includegraphics[width=0.5\textwidth]{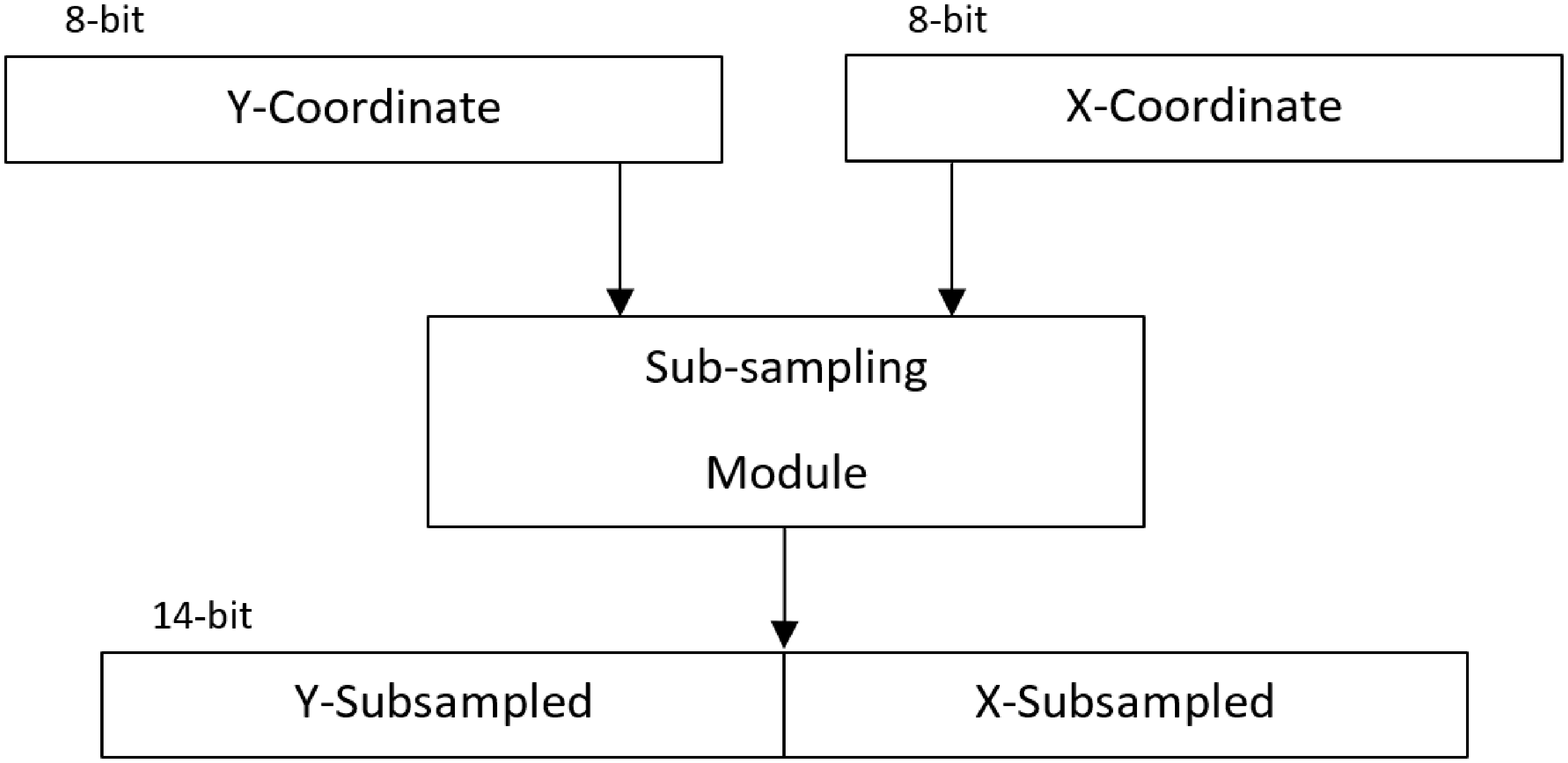}}

\subfloat[A \emph{k}-d tree node in hardware]{\includegraphics[width=0.9\textwidth]{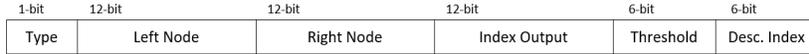}}

\subfloat[Recursive logic-driven \emph{k}-d tree implemented in hardware]{\includegraphics[width=0.7\textwidth]{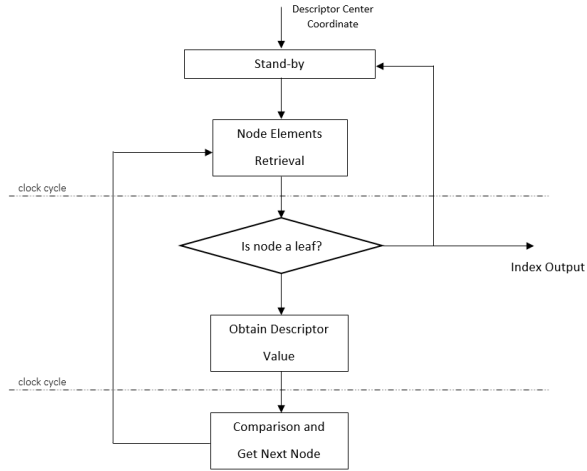}}

\caption{FPGA implementation details}
\label{fig:fpgaimplement}
\end{figure}
\par
The sub-sampling block receives the filtered event locations as input values \emph{x} and \emph{y}, each 8-bit in size, which are used to update the zero-padded count matrix $C \in \mathbb{R}^{m \times n}$ (Eq. \eqref{eq:queue1} and Eq. \eqref{eq:queue2}). The sub-sampling behavior can be achieved in hardware through a combinatorial module that performs the division by shifting the inputs by one bit, and subsequently adding $p$ and $q$ to that value to obtain the sub-sampled representation (Eq. \eqref{equ1}). This results in two 7-bit values which are then concatenated to output a single memory address (Fig.~\ref{fig:fpgaimplement}(a)).
\par
The next block uses the cell-count matrix $R \in \mathbb{R}^{(m/p) \times (n/q)}$, created by a block of distributed RAM of depth $((m/p) \times (n/q))$ and $log(s)$-bits width, corresponding to the FIFO buffer size $s$, initialized to zero for generating the vPCA-RECT representation. To generate a descriptor with respect to the last event received would add a considerable overhead, since each element of the descriptor would have to be read sequentially from the block RAM while being stored by the next module. Instead, the address corresponding to the center of the descriptor is provided, i.e. the input address of the count matrix is passed over to the \emph{k}-d tree module. This allows to trigger the \emph{k}-d tree in one clock cycle once the count matrix is updated and later read the descriptor values based on this single coordinate. However, a new issue arises, the count matrix then can not be modified while the \emph{k}-d tree exploration is being performed. Hence a buffering element is added between the sub-sampling and count matrix modules that will only provide the next address once there is a valid output from the tree.
\par
The \emph{k}-d tree nodes are represented in a 49-bit number stored in a previously initialized single port ROM of depth equal to the number of nodes. This number is conformed by the elements of a node: type, left node, right node, index output, split value and split dimension; these are concatenated and their width is shown in Fig.~\ref{fig:fpgaimplement}(b). 
\par
The \emph{k}-d tree module follows a three steps cycle (Fig.~\ref{fig:fpgaimplement}(c)). The split dimension of a \emph{k}-d tree node provides the address that needs to be read from the cell-count matrix block RAM to get the relevant descriptor value. Next, the descriptor value is compared to the previously stored split value from the node, taking a path down the tree, left or right, depending on the boolean condition. The corresponding node to get is then retrieved from the respective left or right address element acquired in the retrieval step. This cycle repeats until the node type belongs to a leaf, then the leaf node output is made available for the classifier module. It is worth mentioning that in the software implementation of this algorithm, once the descriptor is formed, it is then normalized before being passed to the \emph{k}-d tree. A normalization step in hardware would add a big overhead to the pipeline, disturbing its throughput, and it was removed from the FPGA implementation after verifying that the overall performance was not affected harshly.
\par
At runtime in a software implementation, the classification is performed by a linear combination of the  weights and a feature vector created by the \emph{k}-d tree after a buffer time of $S$ events. To achieve this in a hardware implementation, the depth of the feature vector would have to be transversed while performing several multiplications which would require a considerable amount of multiplier elements from the FPGA, and would affect the speed of the module. Thus, it was desired to avoid this solution and the following was proposed. 
\par
The elements of the linear combination mentioned would be acquired as readily available and would be added to an overall sum vector of length equal to the number of classes to classify, hence performing the dot product operation as one addition per event. Then, after \emph{S} events, a resulting vector is formed, which is equal to the result of the same linear combination first mentioned in the software implementation. Thus, the final module to perform the classification receives the output index from the \emph{k}-d tree and adds its corresponding classifier parameter to a sum vector of length equal to the number of classes. In parallel, this index value is stored in another FIFO element. When the queue is full, the oldest value would be passed to the module to be subtracted from the sum. This allows to have a classification output at any point in time, corresponding to the last \emph{S} events. 
\subsection{Detection Pipeline}
\label{subsec:fpgadetection}
\begin{algorithm}[t]
\caption{Event-based FPGA Object Detection}
\label{algFPGAobjdet}
\textbf{Input}: Filtered event stream $\left\{ \hat{\bf{e}} \right\}$, detector landmarks \emph{l}, number of events \emph{S} \\
\textbf{Output}: Mean object location $(x_{obj},y_{obj})$
\begin{algorithmic}[1]
\State{Initialize detector count $D(y,x) = 0_{m,n}$, detector cut-off $threshold = 0$}
\For{$t=1:S$}
\State{For each incoming event $\hat{\bf{e}}_t =(x_t,y_t,t_t,p_t,{\bf{x}}_t^T)^T$}
\State{For ${\bf{x}}_t$ get leaf node index $l_t$ using \emph{k}-d tree}
\If{$l_t \in \emph{l}$}
\State{$D(y_t,x_t) = D(y_t,x_t) + 1$}
\If{$D(y_t,x_t) > threshold$}
\State{$threshold = threshold + 1$}
\State{Reset detector mean calculation FIFO}
\EndIf
\If{$D(y_t,x_t) = threshold$}
\State{Push $x_t,y_t$ into the mean calculation FIFO}
\EndIf
\EndIf
\EndFor
\State{Output the mean of the coordinates in the FIFO as $(x_{obj},y_{obj})$}
\end{algorithmic}
\end{algorithm} 
Parallel to the modules performing the classification pipeline, the aim of the detection process is to find the coordinates corresponding to ``landmarks" with the highest activation after \emph{S} events, and then find the most probable location for the object. Again, the algorithm was divided into multiple coherent hardware modules that would produce the same results as the original software version. The designed blocks are: landmarks detector, detection heat map and mean calculation.
\par
First, the dictionary features corresponding to the landmarks that were calculated offline are loaded into a binary memory block. This module receives as input the dictionary feature index provided by the \emph{k}-d tree for the current event. If the feature is found as one of the landmarks, the respective event coordinates \emph{x} and \emph{y} are passed as a concatenated address to the next module in the pipeline. Next, a stage corresponding to the heat map is utilized. This module holds a matrix represented as a block RAM of depth ${m \times n}$, since the coordinates are not sub-sampled and have the ranges $1 \leq x \leq m$ and $1 \leq y \leq n$. For each new input address, its value in memory is incremented. 
\par
Since the aim of the detection algorithm is to calculate the average of the coordinates with the highest activation, it would be inefficient to find these event addresses after \emph{S} events. Therefore, the coordinates with the highest count are stored in a FIFO element while the counting is performed. At the end, this will contain all the \emph{x} and \emph{y} coordinates needed for the average calculation. Once the classification flag is triggered, all the coordinates stored in the previous step (which belong to the highest activation) are acquired for calculating the total activation (the divisor). Subsequently, it will calculate the sum of the respective \emph{x} and \emph{y} values, and pass these as dividends to hardware dividers that will provide the final coordinates of the detected object. Alg. \ref{algFPGAobjdet} summarizes the above object detection hardware pipeline clearly.  
\section{Experiments and Discussion}
\label{sec:results}
\subsection{Event-based Object Categorization}
\begin{figure}[t]
	\centering
\subfloat{\includegraphics[width=0.45\textwidth]{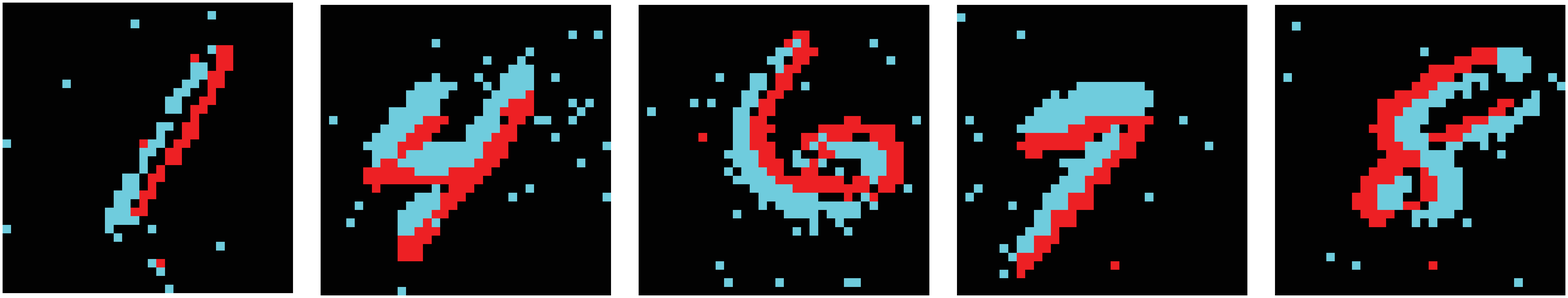}}

\begin{minipage}{9cm}
\center
(a). N-MNIST Samples 
\end{minipage}
\\
\subfloat{\includegraphics[height=0.16\textwidth]{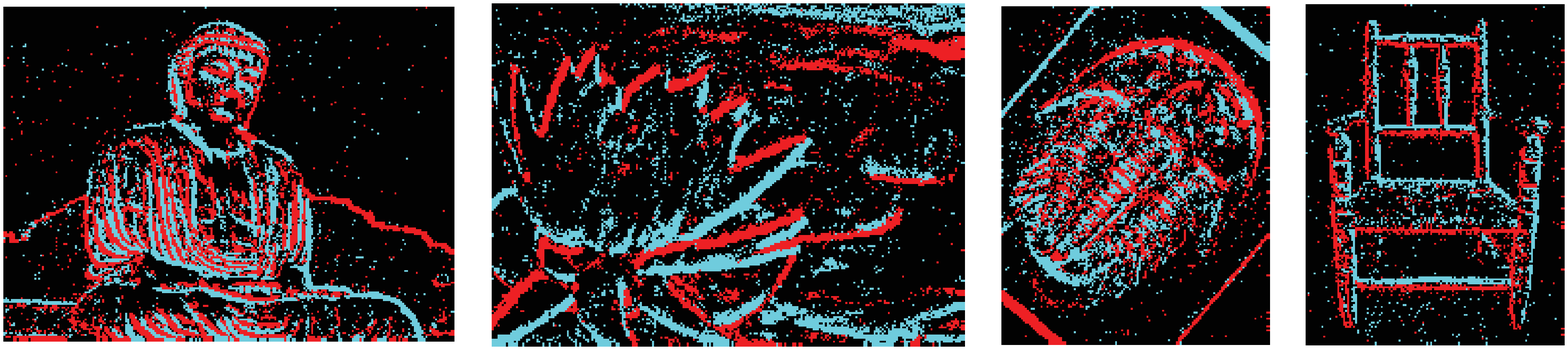}}

\begin{minipage}{9cm}
\center
(b). N-Caltech101 Samples 
\end{minipage}
\caption{Samples from the Event-based Benchmark Datasets}
\label{fig:datasetsamples}
\end{figure}

This section compares the proposed object categorization system to state-of-the-art event-based works and thus software implementation is used with double numeric precision. We validated our approach on two benchmark datasets \cite{Orchard2015}, namely the N-MNIST and N-Caltech101, that have become de-facto standards for testing event-based categorization algorithms. Fig.~\ref{fig:datasetsamples} shows some representative samples from N-MNIST and N-Caltech101.

\subsubsection{Parameter Settings.} 
The time thresholds for the nearest neighbour filter and the refractory filter are nominally set as ${{\bf{\Theta }}_{noise}} = 5$ ms and ${{\rm{\Theta }}_{ref}}= 1$ ms respectively, as suggested in \cite{Padala2018}. We used a FIFO buffer size of 5000 events for dynamically updating the count matrix as and when events are received. Subsequently, the RECT representation with a $2$ by $2$ averaging filter without zero padding at the boundaries is used to obtain a $9\times9$ feature vector for all event locations. We also experimented with other feature vector dimensions using a $3\times3$, $5\times5$, $7\times7$ sampling region and found that increasing the context improved the performance slightly. For obtaining the PCA-RECT representation, the number of PCs can be chosen automatically by retaining the PCs that hold 95\% eigenenergy of the training data, which is typically about 60 in our case. For testing on the benchmark datasets, a dictionary size of 3000 was universally used with a \emph{k}-d tree with backtracking to find precise feature matches. 

\subsubsection{Results on the Benchmark Datasets.} 
The results on the N-MNIST and N-Caltech101 datasets are given in Table~\ref{table:accuracy}. As it is common practice, we report the results in terms of classification accuracy. The baselines methods considered were HATS \cite{Sironi2018}, HOTS \cite{Lagorce2016}, HFirst \cite{Orchard2015a} and Spiking Neural Networks (SNN) \cite{Lee2016,Neil2016a} (Gabor-SNN reported in \cite{Sironi2018}). 
\setlength{\tabcolsep}{4pt}
\begin{table}[t]
\begin{center}
\caption{
Comparison of classification accuracy on event-based datasets (\%).
}
\label{table:accuracy}
\begin{tabular}{lll}
\hline\noalign{\smallskip}
~ & {\bf N-MNIST} & {\bf N-Caltech101} \\
\noalign{\smallskip}
\hline
\noalign{\smallskip}
H-First  & 71.20 & 5.40\\
HOTS & 80.80 & 21.0\\
Gabor-SNN & 83.70 & 19.60\\
HATS & {\bf 99.10} & 64.20\\
vPCA-RECT (this work) &  98.72& 70.25\\
PCA-RECT (this work) & 98.95 & \textbf{72.30}\\
\hline
Phased LSTM & 97.30 & -\\
Deep SNN & 98.70 & -\\
\hline
\end{tabular}
\end{center}
\end{table}
\setlength{\tabcolsep}{1.4pt}
\par
On the widely reported N-MNIST dataset, our method is as good as the best performing HATS method. Moreover, other SNN methods are also in the same ballpark, which is due to the simple texture-less digit event streams giving distinct features for most methods. Therefore, it is a good benchmark as long as a proposed method performs in the high 90's. A test on the challenging NCaltech-101 dataset will pave way for testing the effectiveness close to a real-world scenario. 
\par
\begin{figure}[t]%
\centering
\includegraphics[width=0.55\columnwidth]{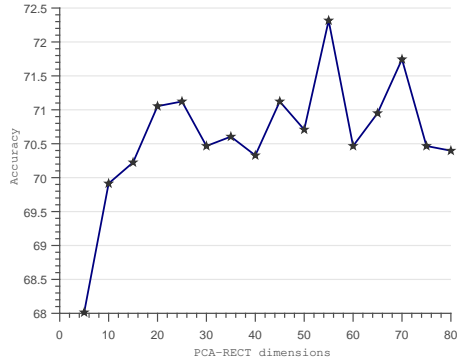}%
\caption{Number of principal component dimensions vs. accuracy on N-Caltech101}%
\label{fig:pcaresults}%
\end{figure}
Our method has the highest classification rate ever reported for an event-based classification method on the challenging N-Caltech101 dataset. The unpublished HATS work is the only comparable method in terms of accuracy, while the other learning mechanisms fail to reach good performance. Fig.~\ref{fig:pcaresults} shows the performance of the PCA-RECT representation as the number of PCs are varied. It is worth noticing that just retaining five dimensions can give better performance compared to available works. 

\subsection{Event-based Object Detection}
\subsubsection{Dataset.}
\begin{figure}[t]
	\centering
\subfloat{\includegraphics[width=0.60\textwidth]{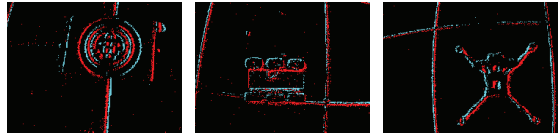}}

\caption{Dataset samples: Landing platform, UAV and Thumper (BG: Empty Floor).}
\label{fig:NSODdatasetsamples}
\end{figure}

The datasets described in the previous section are good for only evaluating the categorization module. In addition, as the benchmark datasets were generated by displaying images on a monitor with limited and predefined motion of the camera, they do not generalize well to real-world situations. To overcome these limitations, we created a new dataset by directly recording objects in lab environment with a freely moving event-based sensor. The in-house dataset, called as Neuromorphic Single Object Dataset (N-SOD), contains three objects with samples of varying length in time (up to 20 s). The three objects to be recognized are a thumper 6-wheel ground robot, an unmanned aerial vehicle, a landing platform along with a background class (Fig.~\ref{fig:NSODdatasetsamples}).

\subsubsection{Results on N-SOD.}
For testing on the N-SOD dataset, we divide the dataset into training and testing, with 80\% temporal sequence samples per class for training and the remaining for testing. Using the training features, a dictionary is generated. Since the temporal sequences are of different length, for a fixed number of events, say every $10^5$ events, a dictionary representation is extracted and a linear SVM classifier is trained. Similarly for testing, for every $10^5$ events, the dictionary representation is classified using the SVM.
\par
Based on the above setup, an accuracy of $97.14\%$ was obtained (Tab.~\ref{tab:confmat}) with a dictionary size of 950, which resulted in a \emph{k}-d tree with 10 layers. We also experimented with lower dictionary sizes such as 150, 300, 450, etc., and the performance drop was insignificant ($> 96\%$). On the other hand, using a \emph{k}-d tree with backtracking, descriptor normalization, etc., achieved close to 100\% accuracy on offline high-performance PCs, which of course does not meet low-power and real-time requirements. In summary, the proposed vPCA-RECT method with a backtracking-free \emph{k}-d tree implementation mildly compromises on accuracy to handle object detection and categorization using an event camera in real-time. 
\begin{table}[t]
\caption{Confusion Matrix (\%) for the best result on the in-house N-SOD dataset.}
\centering
\begin{tabular}{lcccc} 
\hline\hline
~      & Background & LP    & Thumper  &  UAV  \\[0.5ex]
\hline
 Background &   95.4128  &  0.3058  &  3.3639  &  0.9174 \\
 LandingPlatform   & 0  & 99.2268  &  0.5155  &  0.2577\\
 Thumper   &  0  &  1.9257 &  96.9739 &   1.1004\\
 UAV   & 0   &      0  &  3.1884  & 96.8116\\[1ex]
\hline
\end{tabular}
\label{tab:confmat}
\end{table}
\par
We report the precision and recall of the detection results by ascertaining if the mean position of the detected result is within the ground truth bounding box. We obtained: (a) \emph{Precision} - $(498/727) = 0.685$: The percentage of the detections belonging to the object that overlap with the groundtruth (b) \emph{Recall} - $(498/729) = 0.683$: The percentage of correct detections that are retrieved by the system. The number of ``landmarks" were set to 20 in the above experiments while similar results were obtained for values such as five and ten.  
\subsubsection{Comparison to CNN.}
In order to compare to state-of-the-art deep neural networks, we recorded a similar dataset to N-SOD using a frame-based camera and transfer learning via AlexNet classified the object images. With an equivalent train/test split compared to N-SOD, perfect performance can be achieved on the clearly captured test images, however, when we recorded a frame-based dataset under fast motion conditions (motion blur), an accuracy of only $79.20\%$ was obtained. It was clear that the black UAV frame when blurred looks like the black-stripped background and creates much confusion. This confirms the disadvantage of using frame-based cameras to handle unconstrained camera motion. Note that fast camera motion leads to only an increase in data-rate for event-based cameras and has no effect on the output. In fact, recordings of N-SOD have significant amount of such fast motions. 

\subsubsection{FPGA Performance.}
The hardware implementation and performance of the Xilinx Zynq-7020 FPGA running at 100 MHz was evaluated by direct comparison with the results of the algorithm's software version in MATLAB. The time taken for a single event to be classified for the worst possible \emph{k}-d tree path was $550$ nanoseconds. The Zynq was interfaced to a down-looking DAVIS camera, on-board an unmanned aerial vehicle flying under unconstrained lighting scenarios. We recommend viewing our submitted video\footnote{https://youtu.be/h3SgXa47Kjc} that clearly shows the classification/detection process better than still images. 
\par
A summary of utilization of hardware elements can be seen in Tab.~\ref{tab:fpgautil}. Due to our low hardware requirements, the dynamic on-chip power increased only by 0.37W while the base FPGA power consumption was in the order of 3W. As a rough comparison, FPGA-based recognition systems like \cite{Mousouliotis2018,Hikawa2015,Zhai2013}, which present solutions running at equal or lower clock frequencies, consume more power than our implementation. 

\begin{table}[t]
\caption{Hardware utilization report for the FPGA running the proposed modules.}
\centering
\begin{tabular}{lcccc} 
\hline\hline
~      & Utilization & Available &  Utilization \%  \\[0.5ex]
\hline
 LUT &   18238     &    53200   &   34.28\\
 LUTRAM & 12124  &  17400 &     69.68\\
 FF  &  2065 &       106400 &   1.94\\
 BRAM  &  48  &      140 &   34.29\\
 DSP &  4 &       220 &  1.82 \\
 IO & 102    &    200  &   51.00\\[1ex]
\hline
\end{tabular}
\label{tab:fpgautil}
\end{table}

\section{Conclusions}
We have demonstrated object detection and categorization in an energy-efficient manner using event cameras, where the only information that is important for the tasks is how edges move, and the event camera naturally outputs it. The proposed PCA-RECT feature takes advantage of this sparsity to generate a low-dimensional representation. The low-dimensional representation is further exploited for feature matching using a \emph{k}-d tree approach, capable of obtaining the best performance on the challenging Neuromorphic Caltech-101 dataset compared to state-of-the-art works. Most importantly, real-time FPGA implementation was achieved with several careful design considerations, such as a backtracking-free \emph{k}-d tree for dictionary matching, a virtual PCA-RECT representation obtained by analyzing the \emph{k}-d tree partitioning of the feature space, etc. To the best of our knowledge, this is the first work implementing a generic object recognition framework for event cameras on an FPGA device, verified in a lab demo setting under unconstrained motion and lighting setup, thereby demonstrating a high performance over resource ratio.   


%
%
%

\end{document}